\pdfoutput=1

\documentclass[11pt]{article}

\usepackage{acl}

\usepackage{times}
\usepackage{latexsym}

\usepackage[T1]{fontenc}

\usepackage[utf8]{inputenc}

\usepackage{microtype}

\usepackage{stfloats}
\usepackage{multirow}
\usepackage{booktabs}
\usepackage{url}
\usepackage{enumitem}
\usepackage{hyphenat}
\usepackage{color,soul}
\usepackage{combelow}
\definecolor{darkgreen}{RGB}{55,166,84}

\title{MiQA: A Benchmark for Inference on Metaphorical Questions}

\author{
  Iulia-Maria Com\cb{s}a \\
  Google Research, Z{\"u}rich \\
  \small{\texttt{iuliacomsa@google.com}} \\ \And
  Julian Martin Eisenschlos \\
  Google Research, Z{\"u}rich \\
  \small{\texttt{eisenjulian@google.com}} \\ \And
  Srini Narayanan \\
  Google Research, Z{\"u}rich \\
  \small{\texttt{srinin@google.com}} \\
  }

\begin{document}
\maketitle
\begin{abstract}
We propose a benchmark to assess the capability of large language models to reason with conventional metaphors.
Our benchmark combines the previously isolated topics of metaphor detection and commonsense reasoning into a single task that requires a model to make inferences by accurately selecting between the literal and metaphorical register. We examine the performance of state-of-the-art pre-trained models on binary-choice tasks and find a large discrepancy between the performance of small and very large models, going from chance to near-human level. We also analyse the largest model in a generative setting and find that although human performance is approached, careful multiple-shot prompting is required.\footnote{The benchmark is available at \href{https://github.com/google-research/language/tree/master/language/miqa}{https://github.com/google-research/language/tree/master/language/miqa}.}
\end{abstract}

\section{Introduction}

Conceptual metaphor is an ubiquitous cognitive mechanism that allows us to structure and reason about abstract concepts by relating them to experiential domains \citep{lakoff2008metaphors, feldman2008molecule}. In language, metaphors allow human communication and reasoning about abstract ideas using concrete notions learned from sensorimotor, emotional, and other embodied experience~\citep{thibodeau2011}: ``a plan is \textit{solid}''; ``the economy \textit{is stumbling}''; ``I \textit{see} what you mean''.

To illustrate the role of metaphor in abstract reasoning, consider the following metaphorical statement: ``the economy is \emph{stumbling}''. According to conceptual metaphor theory (CMT), we understand this statement through mental simulation, by connecting the abstract concept of economy to the imagined movement of a stumbling person. We use the same mental imagery to infer that the economy, just like a stumbling person, is ``unstable'' and ``might fall''. 
As another example, in the metaphorical statement ``a proposal is \emph{solid}'', bringing to mind solid objects and their properties suggests that the proposal, just like a physical object, was ``well-built'' and ``will not easily break''. 
However, not all properties generalize: unlike a physical object, a proposal, which is an abstract entity, cannot be ``thrown'' or ``bent''; 
unlike a stumbling person, the economy does not ``wear shoes''.

Large Language Models (LLMs) have achieved remarkable results on a variety of tasks. However, in contrast to humans, LLMs do not have access to commonsense and embodied experiences of the world~\citep{bender-koller-2020-climbing}. Although the data LLMs are trained on includes up to trillions of text tokens, it is unclear how much of this data allows them to capture human commonsense reasoning \citep{gordon2013reporting, becker-etal-2021-reconstructing}. Conceptual metaphor theory suggests that embodied and implicit knowledge is required for the ability to utilize metaphors in commonsense reasoning about concrete and abstract ideas.

We propose a novel dataset, MiQA (Metaphorical Inference Questions and Answers), to assess the ability of a model to reason with conventional metaphors.
The benchmark draws on the CMT research \citep{grady1997foundations} to construct a representative set of primary metaphors, which are contrastively paired with literal statements.   
Our task requires a model to make a correct inference in simple situations without specifying whether the contextual register is literal or metaphorical, leveraging research on metaphor processing \citep{rai2020}, commonsense reasoning \citep{davis2015}, and natural language inference \citep{dagan2006, bowman-etal-2015-large}.

Our benchmark combines the previously isolated areas of metaphor detection and commonsense inference. Although there is considerable research on both of these areas separately, it is unclear whether such capabilities scale compositionally: LLMs could handle two separate tasks well, but not their combination~\citep{Keysers2019}.

Our contributions are the following:
\begin{itemize}[nolistsep]
    \setlength\itemsep{0.25em}
    \item We propose MiQA, a benchmark for commonsense inference with conventional metaphors;
    \item We show a large discrepancy between the performance of small and large models in a binary-choice MiQA task, from chance to human-level accuracy;
    \item We use a generative MiQA task to corroborate the performance of the largest model in an open-ended setting, showing that although human-level performance is approached, careful multiple-shot prompting is required. 
\end{itemize}

\begin{table*}
\centering
\small
\begin{tabular}{p{7.25cm}p{7.25cm}}
\toprule
\textbf{``\textit{implies}''-questions} & \textbf{``\textit{implied-by}''-questions}\\
\midrule
``I see what you mean''. \newline 
Which of the following statements could that imply? \newline
(1) My eyes are working well \textcolor{red}{\textbf{[incorrect]}} \newline
(2) \underline{I understand you} \textcolor{darkgreen}{\textbf{[correct]}}
&
``My eyes are working well'' \newline
is implied by which of the following? \newline 
(1) \underline{I see what you are pointing at} \textcolor{darkgreen}{\textbf{[correct]}} \newline
(2) I see what you mean \textcolor{red}{\textbf{[incorrect]}}
\\
\midrule
``A plan is not solid''. \newline 
Which of the following statements could that imply? \newline
(1) A hammer could break it \textcolor{red}{\textbf{[incorrect]}} \newline
(2) \underline{We should not follow it} \textcolor{darkgreen}{\textbf{[correct]}}
&
``A hammer could break it'' \newline
is implied by which of the following? \newline 
(1) \underline{A table is not solid} \textcolor{darkgreen}{ \textbf{[correct]}} \newline
(2) A plan is not solid  \textcolor{red}{\textbf{[incorrect]}}
\\
\midrule
``My friend has a huge problem''. \newline 
Which of the following statements could that imply? \newline
(1) My friend needs space \textcolor{red}{\textbf{[incorrect]}} \newline
(2) \underline{My friend needs a solution} \textcolor{darkgreen}{\textbf{[correct]}}
&
``My friend needs space'' \newline
is implied by which of the following? \newline 
(1) \underline{My friend has a huge dog} \textcolor{darkgreen}{\textbf{[correct]}} \newline
(2) My friend has a huge problem \textcolor{red}{\textbf{[incorrect]}}
\\
\bottomrule
\end{tabular}
\caption{\label{miq-questions}
Examples of MiQA items combined into two task types. Correct answers are labelled. The examples are built by pairing $150$ sets of two premises (one metaphorical and one literal) and two respective implications.
}
\end{table*}

\section{Related Work}

Metaphor has received renewed attention in natural language processing, but most tasks have been restricted to detection (e.g. \citealp{shared-metaphor-task-2018, shared-metaphor-task-2020, choi-etal-2021-melbert}) on large annotated corpora \citep{steen2010method, beigman-klebanov-etal-2016-semantic}. Human-level performance has not been reached by LLMs, but the progress is promising and an active area of research. However,
these tasks may excessively rely on context-dependent word meanings \citep{neidlein-etal-2020-analysis} and do not measure the ability to reason with metaphor.

Metaphor paraphrasing is another active area of research. BIG-bench \citep{big-bench}, a collaborative multi-task benchmark intended to test a variety of capabilities of LLMs, includes four tasks related to metaphor.
While these tasks contain novel metaphors, they still do not assess the ability to employ metaphoric knowledge in reasoning.

Recently, \citet{chakrabarty-etal-2022-rocket} built a dataset for multiple-choice story continuations involving similes and idioms extracted from books. 
Subsequently, \citet{fig-qa-paper} proposed a metaphor interpretation task that requires models to choose the correct out of two interpretations of a simile. 
In contrast, our task combines metaphor interpretation with commonsense inference, uses a more systematic data source, and has an additional adversarial character, as it requires the selection between two semantically-close items instead of items with opposite meanings.

\section{Dataset}
\label{dataset}

\subsection{Motivation}

Most existing studies of metaphor have primarily started from corpus-based methods, using frequency and other corpus-based metrics to detect or classify metaphors. This process leads to a primary focus on corpus distributions and makes it hard to compare studies across different corpora. Furthermore, it ignores the central tenet of metaphor theory that foundational metaphors are grounded in non-linguistic and experiential domains which may be assumed as a background and thus underrepresented in corpora. 

To address this, we chose to use a foundational ontology of primary conceptual metaphors~\citep{grady1997foundations} based on CMT~\citep{lakoff2008metaphors, feldman2008molecule}. Our choice of primary metaphors has multiple desirable properties. Primary metaphors are a good starting point for the investigation of more complex, compositional metaphor. The metaphors in our dataset are developmentally early in child experience with primary scenes and language. Moreover, the chosen metaphors are embodied, in that the source domain is often observable and sensory-motor (size, warmth, height) while the targets are less observable and often subjective or abstract (importance, affection, quantity). The metaphors we chose form a basis set of mappings that can create, through composition, more complex mappings, such as the Event Structure Metaphor, that maps movement and manipulation to actions. Our approach ensures that the distributions of metaphor categories in the task is balanced and hence reflective of the capability of large models to use primary metaphors as building blocks of reasoning.

\subsection{Construction}

We constructed a novel dataset consisting of $150$ items. 
The items were manually created by the authors based on the work of \citet{grady1997foundations}, which lists $100$ primary metaphors that are conventional, developmentally early, and form a basis set for composing complex mappings.

Each item in the dataset is a tuple consisting of four sentences in English: a literal premise ($L_p$), a premise containing a conventional metaphor ($M_p$), an implication of the literal premise ($L_c$), and an implication of the metaphorical premise ($M_c$).

The tuples are paired so that a mistaken literal interpretation of $M_p$ can falsely suggest that $L_c$ is implied. For example, a wrong inference would be:
``\textit{I see what you mean}'' implies that ``\textit{I am using my eyes}''.
The false implication $M_p \to L_c$ thus serves as an adversarial element that probes whether the model correctly registers the metaphorical context.

For each primary metaphor proposed by \citet{grady1997foundations}, we manually created $1$-$2$ pairs of items in the form described above, where $M_p$ is an example of the metaphor, while $L_p$ relates to the source domain of the metaphor only. 

To create the final benchmark for LLMs, we used these items to generate two types of adversarial questions. 
The first type (``\textit{implies}''-questions) requests the model to select the most likely inference given a metaphorical statement. Answering correctly requires the model to not be tricked by a possible literal interpretation of the metaphorical premise.
The second type (``\textit{implied-by}''-questions) requests the model to select the most likely premise that a literal statement is implied by. Answering correctly requires the model to not be tricked by a possible metaphorical interpretation of the literal conclusion.
See Table~\ref{miq-questions} for examples.


We combined these items to obtain a benchmark consisting of $300$ questions, of which half are ``\textit{implies}''-questions and half are ``\textit{implied-by}''-questions. This pairing of tasks ensures that the model does not achieve a better score if biased towards assigning a higher likelihood to either literal or metaphorical continuations of a statement.

\section{Human Evaluation}

We estimated the human performance on the binary-choice task using the responses of $15$ human adult volunteers with English as first or second language.
The participants were told that the aim of the research was to gather a set of commonsense responses and compare them to LLMs responses. No additional information about the task was given.

\section{Large Language Models Evaluation}

\begin{table*}
\centering
\begin{tabular}{p{15cm}}
\toprule

\textbf{Prompt:} ``I am sticking to my plans''. Could that imply that ``I am using an adhesive''? \\
\textbf{Scored as \textcolor{darkgreen}{correct}:} No, it means that you are not changing your plans. \\
\midrule

\textbf{Prompt:} `"I have to move ahead with my project". Could that imply that "a trolley could help"? \\
\textbf{Scored as \textcolor{red}{wrong}:} I think it's a reference to the trolley problem. \\
\midrule

\textbf{Prompt:} ``My future looks bright''. Could that imply that ``I could use sunglasses''? \\
\textbf{Scored as \textcolor{gray}{ambiguous}:} I think it's a bit of a stretch. \\

\bottomrule
\end{tabular}
\caption{\label{palm-generative}
Examples of prompts from the MiQA dataset and manual scoring of answers generated by PaLM-540b.
}
\end{table*}

We evaluated the performance of two pre-trained LLMs: 
PaLM with 8B, 62B, 540B parameters \citep{palm-paper} and GPT-3 Ada, Babbage, Curie and DaVinci \citep{gpt3-paper}. The parameter counts of the GPT-3 models are not publicly available, but have been estimated at 350M, 1.3B, 6.7B, and 175B respectively \footnote{\href{ https://blog.eleuther.ai/gpt3-model-sizes}{https://blog.eleuther.ai/gpt3-model-sizes}}.

The main purpose of this study is to assess the capabilities of LLMs on the MiQA benchmark. For comparison, we also verify the capabilities of pre-trained fine-tuned smaller language models on our benchmark. We follow~\citet{fig-qa-paper} in using encoder-only models trained on the natural language inference datasets SNLI~\citep{bowman-etal-2015-large} and MNLI~\citep{williams-etal-2018-broad} for zero-shot evaluation.
We opt for this approach because MiQA is designed as a small dataset suitable as a benchmark and not for fine-tuning.
We test the state-of-the-art encoder-only model DeBERTaV3~\citep{he2021debertav3} in sizes small, medium and large. These models have 44M, 86M and 304M parameters respectively, and their weights are available online\footnote{\href{https://huggingface.co/models?search=cross-encoder/nli-deberta-v3}{https://huggingface.co/models?search=cross-encoder/nli-deberta-v3}}.
The models take a premise-implication pair and produce a probability distribution over three classes: ``entailment'', ``contradiction'', and ``undetermined''. We report the results for the best-performing score, in this case $1 - P(\emph{contradiction})$, which outranks $P(\emph{entailment})$.

\subsection{Binary-Choice Tasks}

We first assessed the models by prompting them with the question types illustrated in Table~\ref{miq-questions} in $0$, $1$ and $5$-shot settings. Few-shot prompts were obtained by prefixing with randomly selected questions followed by their correct answers. To score the results, we obtained the log likelihood of each of the two choices as candidate continuations to the given prompt question. A response was scored as correct if the log likelihood of the correct choice was larger than that of the incorrect choice.

Prompts can greatly influence LLM predictions \citep{lu-etal-2022-fantastically, cao-etal-2021-knowledgeable}. As expected, we observed variability with changing prompts. To mitigate this, we tried multiple prompts, as detailed in Appendix~\ref{appendix}. For each model, we selected the prompt that performed best with $0$-shot, and subsequently used this prompt to obtain and report its results in few-shot settings. Additionally, we used two baseline prompts (an empty prompt with no choices, and a prompt containing an unrelated question), which can indicate if the models simply learn to select either metaphorical or literal statements independently of the prompt in few-shot settings.
 
\subsection{Generative Task}

In addition to the binary-choice tasks, we also tested the largest model, PaLM-540b, in a generative setting. We prompted this model with questions of the form: ``$M_p$. Could that imply that $L_c$?''. This capitalises on the adversarial false implication $M_p \to L_c$ described in section~\ref{dataset}. 

We obtained completions to the $150$ questions of this form generated from the MiQA dataset. Answers were manually and independently scored by every author. Every author scored at least two thirds of all responses and the scores were averaged. Scoring consisted of labelling the first paragraph of an answer as ``correct'', ``wrong'' or ``ambiguous''. To compute the accuracy over the generative task, ``correct'' responses were scored as $1$ and ``ambiguous'' responses were scored as $0.5$. Agreement between raters was medium, with intraclass correlation \citep{shrout1979intraclass} $ICC(2, k)$ at $0.56$. Examples of scored answers are shown in Table~\ref{palm-generative}. 

As before, we evaluated the model in $0$, $1$ and $5$-shot settings. From the $0$-shot setting, we selected $32$ answers produced by the model that all authors independently scored as ``correct'', and the same number of answers of the form $M_p \to M_c$ and $L_p \to L_c$. We randomly selected $1$ or $5$ of these answers and their corresponding questions to prefix each prompt question in $1$- and $5$-shot settings.

\section{Results} 

\begin{table}[t!]
\footnotesize
\centering
\begin{tabular}{lccc}
\toprule
\multicolumn{1}{c}{} & \multicolumn{3}{c}{\textbf{Accuracy}} \\
\midrule
\midrule
 & \multicolumn{3}{c}{\textbf{``\textit{implies}'' questions}} \\ 
\cmidrule{2-4}

\textbf{Model\quad\quad\quad Shots} & \textbf{\quad0\quad} & \textbf{\quad1\quad} & \textbf{5} \\

\midrule
PaLM-8b                                 & 54.0  & 51.3  & 51.0          \\
PaLM-62b                                & 53.7  & 62.0  & 66.3          \\
PaLM-540b                               & \textbf{89.7}  & \textbf{97.0}  & \textbf{96.3}           \\
\midrule
GPT-3-Ada                               & 52.3  & 49.0  & 51.3           \\
GPT-3-Babbage                           & 50.7  & 51.0  & 51.3           \\
GPT-3-Curie                             & 50.7  & 57.3  & 55.7           \\
GPT-3-DaVinci                           & \textbf{89.3}  & \textbf{97.7}  & \textbf{98.7}           \\
\midrule
DeBERTaV3-NLI-small                     & 78.0 & & \\
DeBERTaV3-NLI-base                      & \textbf{82.7} & & \\
DeBERTaV3-NLI-large                     & 80.0 & & \\
\midrule
\multicolumn{1}{l}{\textbf{Baseline prompts}} & & & \\ 
no choices                 & 48.7  & 49.3  & 52.3           \\
no questions             & 50.0  & 56.0  & 58.0            \\
\midrule
Human                                   & \textbf{99.6}  &   &    \\

\midrule
\midrule

 & \multicolumn{3}{c}{\textbf{``\textit{implied-by}'' questions}} \\ 
\cmidrule{2-4}
\textbf{Model\quad\quad\quad Shots} & \textbf{0} & \textbf{1} & \textbf{5} \\
\midrule
PaLM-8b                                 & 51.0   & 57.7     & 55.3  \\
PaLM-62b                                & 53.0   & 58.0     & 65.3  \\
PaLM-540b                               & \textbf{71.3}   & \textbf{84.7}     & \textbf{92.3}  \\
\midrule
GPT-3-Ada                               & 53.0   & 53.0     & 46.3  \\
GPT-3-Babbage                           & 50.7   & 50.3     & 52.0  \\
GPT-3-Curie                             & 55.0   & 53.7     & 50.0  \\
GPT-3-DaVinci                           & \textbf{77.3}   & \textbf{88.0}     & \textbf{95.7}  \\
\midrule
DeBERTaV3-NLI-small                     & 74.0 & & \\
DeBERTaV3-NLI-base                      & 70.7 & & \\
DeBERTaV3-NLI-large                     & \textbf{76.7} & & \\
\midrule
\multicolumn{1}{l}{\textbf{Baseline prompts}} & & & \\ 
no choices                 & 49.3   & 53.0     & 58.7  \\              
no questions             & 50.0   & 43.0     & 47.0  \\
\midrule

Human                                   & \textbf{96.4}   &       &    \\
\midrule
\midrule
Chance                                  & 50.0  &        &    \\

\midrule
\midrule

\textbf{Generative task}                & \textbf{71.7}    &  \textbf{73.1}  & \textbf{88.9} \\
\bottomrule

\end{tabular}
\caption{\label{miq-results}
Results obtained on the MiQA tasks by pre-trained LLMs in few-shot settings.
Small models perform close to chance level, while large models perform close to human level.  We include two prompting baselines for PaLM-540b, whose performance close to chance level shows that few-shot performance is not due to metaphor detection only. We also include the accuracy for the generative task that asks PaLM-540 to answer open-ended ``$M_p \to L_c$?'' questions, scored as descried in Table~\ref{palm-generative}. Multiple-shot prompting is required to approach human-level performance. 
}
\end{table}

The full results are shown in Table~\ref{miq-results}.

Firstly, for the binary-choice tasks, there was a considerable gap between small and large LLMs. While the smaller models performed at or close to chance level, the largest models achieved very good performance even with $0$ shots, and approached human-level performance with few shots. We note that the ``\textit{implied-by}'' task was overall more difficult than the ``\textit{implies}'' task for both humans and LLMs. 

Secondly, the chance-level performance on the baseline prompts suggests that the increase in performance in few-shot settings was not due to the model learning to select either metaphorical or literal statements independently of the prompt. 
On the other hand, the strong performance of the DeBERTaV3 models suggests a high level of transfer from the NLI datasets to MiQA, although there is a still a considerable gap to human performance.

Finally, the generative results on PaLM-540b estimated the model performance in an open-ended setting. Similarly to the binary task, the model performed considerably better with $5$ shots compared to $0$ shots, approaching human performance. However, the gap between human and model performance for the generative task was greater compared with the gap for the binary-choice task.

Overall, the results demonstrate that LLMs can correctly select between the metaphorical and literal contextual registers to perform inferences with conventional metaphors, but there is still a considerable gap between human and LLM performance in $0$-shot settings.

\section{Limitations}

Our work used foundational metaphors from CMT to test basic metaphoric reasoning in LLMs.
We will expand this benchmark using additional and more complex sources of conceptual metaphor (e.g. \citealp{metanet}). Future work will assess LLMs on novel non-conventional mappings.

Although we mitigated for prompt sensitivity by using multiple prompts, the result interpretation should allow for small accuracy variations. 
Further, in the binary-choice tasks we compare the LLM results with a human baseline, but we do not provide a baseline for the generative task. This is near perfect for humans, but a more systematic baseline can be created to quantify the exact headroom on this task. 
Finally, while the task holistically measures the performance of LLMs on a complex task, it is difficult to disentangle the component effects 
(metaphor detection, reasoning, response generation) in the overall accuracy. 

\section{Conclusion}

We have proposed a novel compositional benchmark based on conceptual metaphor theory to assess the capacity of LLMs to make inferences with metaphors. Successful performance on this task requires metaphor detection and commonsense inference. Using a metaphor theory-based approach allows us to systematically explore capabilities and limitations of LLMs. This is the first in a planned series of increasingly complex metaphor inference datasets.

Three main findings emerged from our proposed task. Firstly, there is a vast difference between the performance of small and large LLMs, with the former performing at chance level and the latter approaching human level in few-prompt setting. This observation is informative in the context of previous results showing that some, but not all, tasks observe a qualitative performance jump with model size and scale: for example, this is the case for reasoning about goal-step relationships between events and ordering events, but not for navigation and mathematical induction tasks~\citep{palm-paper}. This result invites more research into the question of how and whether the performance of smaller models can be improved. Secondly, this reflects a true ability of LLMs to reason with conventional metaphor, and not simply to detect it. Whether this ability extends to novel metaphor is ongoing work. Finally, the performance of large LLMs approaches that of humans in binary-choice and generative tasks, but careful multiple-shot prompting is required.

\newpage

\section*{Acknowledgements}

We thank Fernando Pereira, Yasemin Altun, \mbox{William} Cohen and Tiago Pimentel, as well as our anonymous reviewers, for their valuable feedback.

\bibliography{anthology,custom}
\bibliographystyle{acl_natbib}

\newpage

\appendix

\section{Appendix}
\label{appendix}

The following prompts were used to assess model performance on the binary-choice MiQA tasks with ``\textit{implies}''-questions:

\begin{enumerate}
  \item \texttt{``$M_p$''. Which of the following two statements could that imply? $L_c$ or $M_c$?} \newline (chosen for PaLM-8b, PaLM-16b, GPT3-Ada, GPT3-Babbage, GPT3-Curie)
  \item \texttt{``$M_p$''. Which of the following two statements could that imply? (1) $L_c$ (2) $M_c$}
  \item \texttt{Q: ``$M_p$''. Which of the following two statements could that imply? (1) $L_c$ (2) $M_c$ A:} \newline (chosen for PaLM-540b)
  \item \texttt{Question: ``$M_p$''. Which of the following two statements could that imply? (1) $L_c$ (2) $M_c$ Answer: It could imply} \newline (chosen for GPT3-DaVinci)
\end{enumerate}

Similar prompts were used with ``\textit{implied-by}''-questions.

The following prompts were used as a baseline. In a $0$-shot setting, these will produce an accuracy related to the independent log likelihood of each candidate answer. In few-shot settings, these will produce a better performance if the model learns to act as a metaphor detector, independently of any statement connecting the two answers:

\begin{enumerate}
  \item \texttt{``''} (empty prompt)
  \item \texttt{Pick between the following statements: (1) $L_c$ (2) $M_c$} (random choice)
\end{enumerate}

We verified that the results were similar when the random choice baseline was altered to be more similar to the best-scoring prompt for the largest model (PaLM-540b).

To minimise the influence of the answer order on the scores of the model, we presented each question twice, swapping the order of the choices the second time. This has the effect of producing a better mean approximation than presenting each question once with randomised order of choices.

\end{document}